\documentclass[10pt,twocolumn,letterpaper]{article}

\usepackage{iccv}
\usepackage{times}
\usepackage{epsfig}
\usepackage{graphicx}
\usepackage{amsmath}
\usepackage{amssymb}
\usepackage{pmat}
\usepackage[noend]{algpseudocode}
\usepackage{algorithmicx,algorithm}
\usepackage{multirow}
\usepackage{array}
\usepackage{color}
\usepackage{booktabs}
\usepackage[misc]{ifsym}
\usepackage{makecell}
\usepackage[breaklinks=true,bookmarks=false,colorlinks]{hyperref}
\newcolumntype{M}[1]{>{\centering\arraybackslash}m{#1}}
\iccvfinalcopy 

\def\iccvPaperID{****} 

\ificcvfinal\fi

\begin{document}

\title{Geometry Uncertainty Projection Network for Monocular 3D Object Detection}
\begin{small}
\author{Yan Lu$^{1,\dagger,*}$\quad Xinzhu Ma$^{1,*}$\quad Lei Yang$^2$\quad Tianzhu Zhang$^3$\\ Yating Liu$^4$\quad Qi Chu$^{3,\text{\Letter}}$\quad 
Junjie Yan$^2$\quad Wanli Ouyang$^{1,\text{\Letter}}$\\
$^1$The University of Sydney, SenseTime Computer Vision Group\quad $^2$Sensetime Group Limited\\
$^3$School of Information Science and Technology, University of Science and Technology of China\\
$^4$School of Data Science, University of Science and Technology of China\\
{\tt\small \{yan.lu1, xinzhu.ma, wanli.ouyang\}@sydney.edu.au\quad \{yanglei, yanjunjie\}@sensetime.com}\\
{\tt\small \{tzzhang, qchu\}@ustc.edu.cn\quad liuyat@mail.ustc.edu.cn}
}
\end{small}
\maketitle
\renewcommand{\thefootnote}{$\dagger$}
\footnotetext{This work was done when Yan Lu was an intern at SenseTime.}
\renewcommand{\thefootnote}{*}
\footnotetext{Equal contribution.}
\renewcommand{\thefootnote}{\Letter}
\footnotetext{Corresponding author.}
\renewcommand{\thefootnote}{}
\footnotetext{\href{https://github.com/SuperMHP/GUPNet/blob/main/pdf/supp.pdf}{Supplementary material link}}
\renewcommand{\thefootnote}{1}

\ificcvfinal\thispagestyle{empty}\fi
\begin{abstract}
Geometry Projection is a powerful depth estimation method in monocular 3D object detection. It estimates depth dependent on heights, which introduces mathematical priors into the deep model. But projection process also introduces the error amplification problem, in which the error of the estimated height will be amplified and reflected greatly at the output depth. This property leads to uncontrollable depth inferences and also damages the training efficiency. In this paper, we propose a Geometry Uncertainty Projection Network (GUP Net) to tackle the error amplification problem at both inference and training stages. Specifically, a GUP module is proposed to obtains the geometry-guided uncertainty of the inferred depth, which not only provides high reliable confidence for each depth but also benefits depth learning. Furthermore, at the training stage, we propose a Hierarchical Task Learning strategy to reduce the instability caused by error amplification. This learning algorithm monitors the learning situation of each task by a proposed indicator and adaptively assigns the proper loss weights for different tasks according to their pre-tasks situation. Based on that, each task starts learning only when its pre-tasks are learned well, which can significantly improve the stability and efficiency of the training process. Extensive experiments demonstrate the effectiveness of the proposed method. The overall model can infer more reliable object depth than existing methods and outperforms the state-of-the-art image-based monocular 3D detectors by 3.74\% and 4.7\% $\text{AP}_{40}$ of the car and pedestrian categories on the KITTI benchmark. The code and model will be released at \href{https://github.com/SuperMHP/GUPNet}{https://github.com/SuperMHP/GUPNet}.
\end{abstract}

\begin{figure}[t]
\begin{center}
\includegraphics[width=1.0\linewidth]{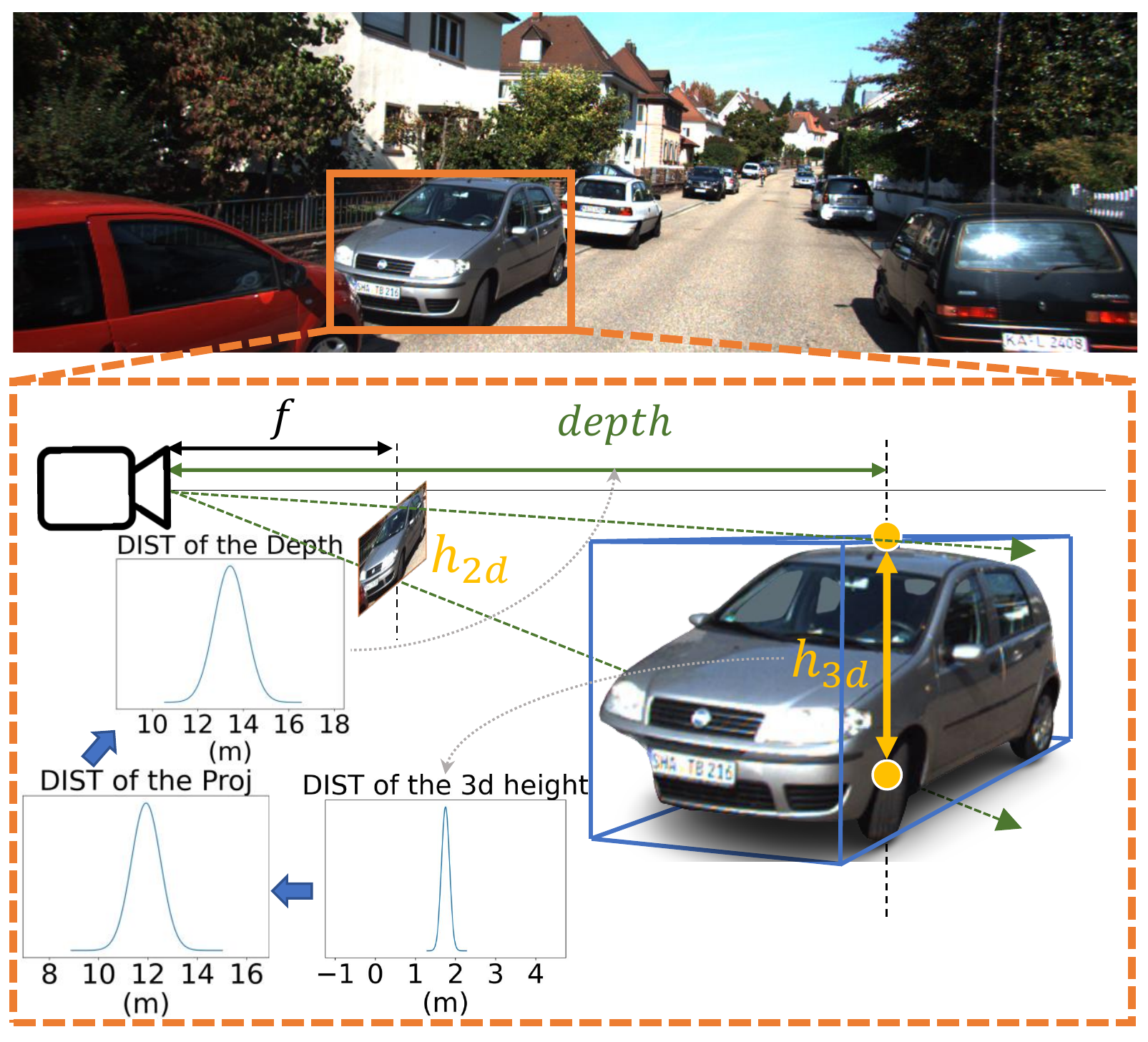}
\end{center}
   \caption{The main pipeline of our Geometry Uncertainty Projection module. The projection process is modeled by the uncertainty theory in the probability framework. The inference depths can be represented as a distribution so that can provide both accurate values and scores.}
\label{fig:upm}
\end{figure}

\section{Introduction}

3D object detection is an important component in autonomous driving and has received increasing attention in recent years. 
Compared with the LiDAR/stereo-based methods~\cite{meyer2019lasernet,qi2019deep,qin2019triangulation,shi2019pointrcnn,shi2019part,yang2019std,yuan2021temporal}, monocular 3D object detection is still a challenging task due to the lack of depth cues, which makes monocular object-level depth estimation naturally ill-posed. Therefore, the monocular 3D detector cannot achieve satisfactory performance even some complex network structures~\cite{roddick2018orthographic} are applied. Recently, to alleviate this problem, some works~\cite{qin2019monogrnet,weng2019monocular} attempt to introduce geometry priors to help depth inference, of which a widely used prior is the perspective projection model. 

\begin{figure}[t]
\begin{center}
\includegraphics[width=1.0\linewidth]{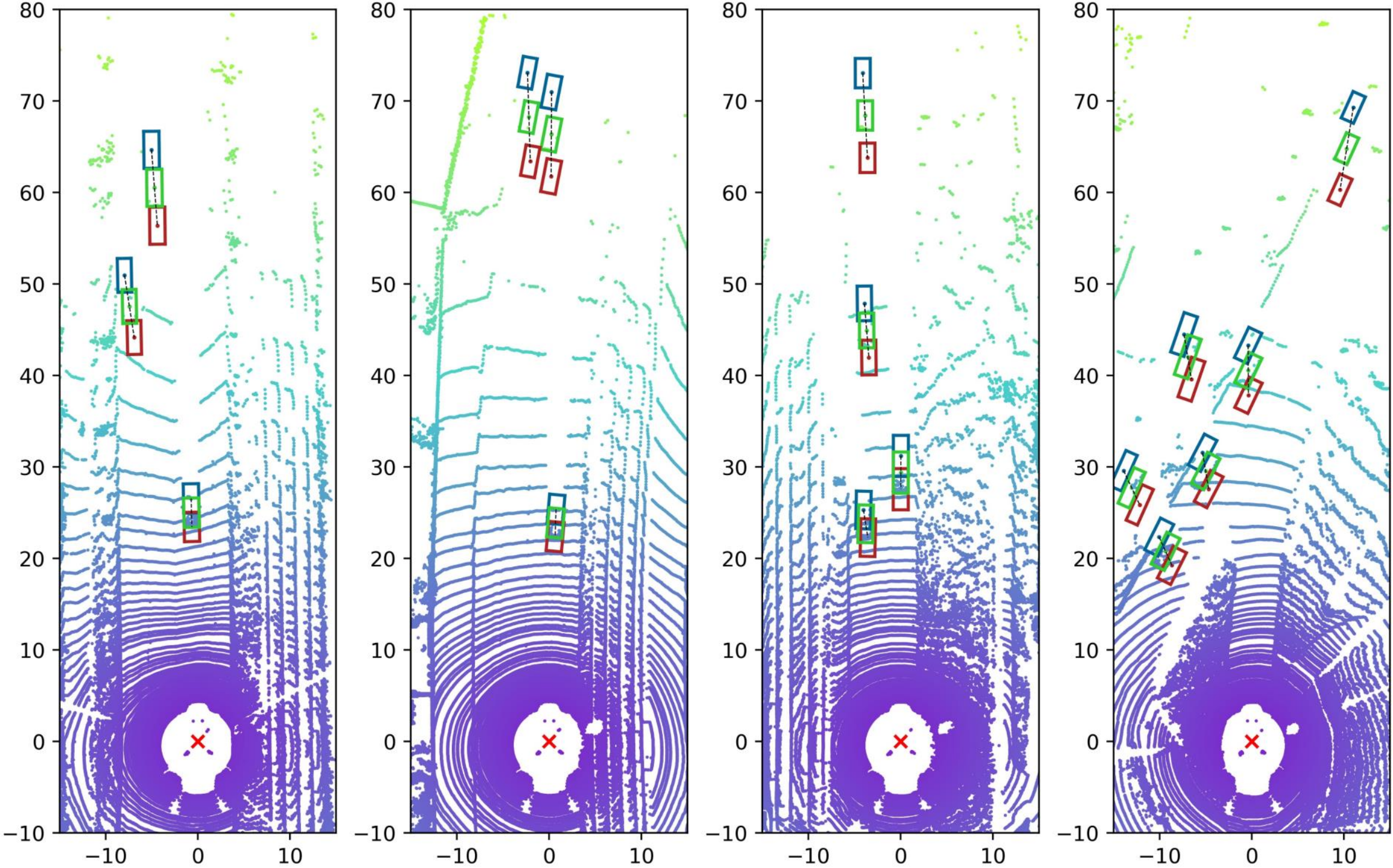}
\end{center}
   \caption{Visualized examples of the depth shift caused by $\pm$0.1m 3D height jitter. We draw some bird's view examples to show the error amplification effect. In this figure, the unit of the horizontal axis and the vertical axis are both meters, and the vertical axis corresponds to the depth direction. The \textcolor[RGB]{50,205,50}{green} boxes mean the original projection outputs. The \textcolor[RGB]{0,102,153}{
  blue} and \textcolor[RGB]{178,34,34}{red} boxes are shifted boxes caused by +0.1m and -0.1m 3D height bias respectively (best viewed in color). }
\label{fig:shift}
\end{figure}

Existing methods with the projection model usually estimate the height of 2D and 3D bounding box first and then infer the depth via the projection formula $depth=h_{3d}\cdot f/h_{2d}$ ($f$ is the camera focal length). Depth inferred by this formula is highly related to the estimated 2D/3D heights so the error of the height estimation will also be reflected at the estimated depth. However, the error of height estimation is inevitable especially for the ill-posed 3D height estimation (2D height estimation is relatively more accurate because of the well-developed 2d detection), so we are more concerned about the depth inference error caused by the 3D height estimation error. To show the influence of this property, we visualize the depth shifts caused by a fixed 3D height error in Figure~\ref{fig:shift}.
We can find that a slight bias (0.1m) of 3D heights could cause a significant shift (even 4m) in the projected depth.
This error amplification effect makes outputs of the projection-based methods hardly controllable, significantly affecting both inference reliability and training efficiency. In this paper, we propose a Geometry Uncertainty Projection Network that includes a Geometry Uncertainty Projection (GUP) module and a Hierarchical Task Learning (HTL) strategy to treat these problems.

The first problem is inference reliability. A small quality change in the 3D height estimation would cause a large change in the depth estimation quality. This makes the model cannot predict reliable uncertainty or confidence easily, leading to uncontrollable outputs. To tackle this problem, the GUP module is proposed to infer the depth based on the distribution form rather than a discrete value (see Figure~\ref{fig:upm}). The depth distribution is inferred by the estimated 3D height distribution. So, the statistical characteristics of the estimated 3D height estimation would be reflected in the output depth distribution, which leads to more accurate Uncertainty. At the inference, this well-learned uncertainty would be mapped to a confidence value to indicate the depth inference quality, which makes the total projection process more reliable.

Another problem is the instability of model training.
In particular, at the beginning of the training phase, the estimation of 2D/3D height tends to be noisy, and the errors will be amplified and cause outrageous depth estimation.
Consequently, the training process of the network will be misled, which will lead to the degradation of the final performance.
To solve the instability of the training, we propose the Hierarchical Task Learning (HTL) strategy, aiming to ensure that each task is trained only when all pre-tasks (\eg 3D height estimation is one of the pre-tasks of depth estimation) are trained well. 
To achieve that, the HTL first measures the learning situation of each task by a well-designed learning situation indicator.
Then it adjusts weights for each loss term automatically by the learning situation of their pre-tasks, which can significantly improve the training stability, thereby boosting the final performance.

In summary, the key contributions of this paper are as follows:
\begin{itemize}
    \item We propose a Geometry Uncertainty Projection (GUP) module combining both mathematical priors and uncertainty modeling, which significantly reduces the uncontrollable effect caused by the error amplification at the inference. 
    \item For the training instability caused by task dependency in geometry-based methods, we propose a Hierarchical Task Learning (HTL) strategy, which can significantly improve the training efficiency.
    \item Evaluation on the challenging KITTI dataset shows the overall proposed GUP Net achieves state-of-the-art performance around 20.11\% and 14.72\% on the car and the pedestrian 3D detection respectively on the KITTI testing set.
\end{itemize}

\section{Related works}
\begin{figure*}[t]
\begin{center}
\includegraphics[width=0.8\linewidth]{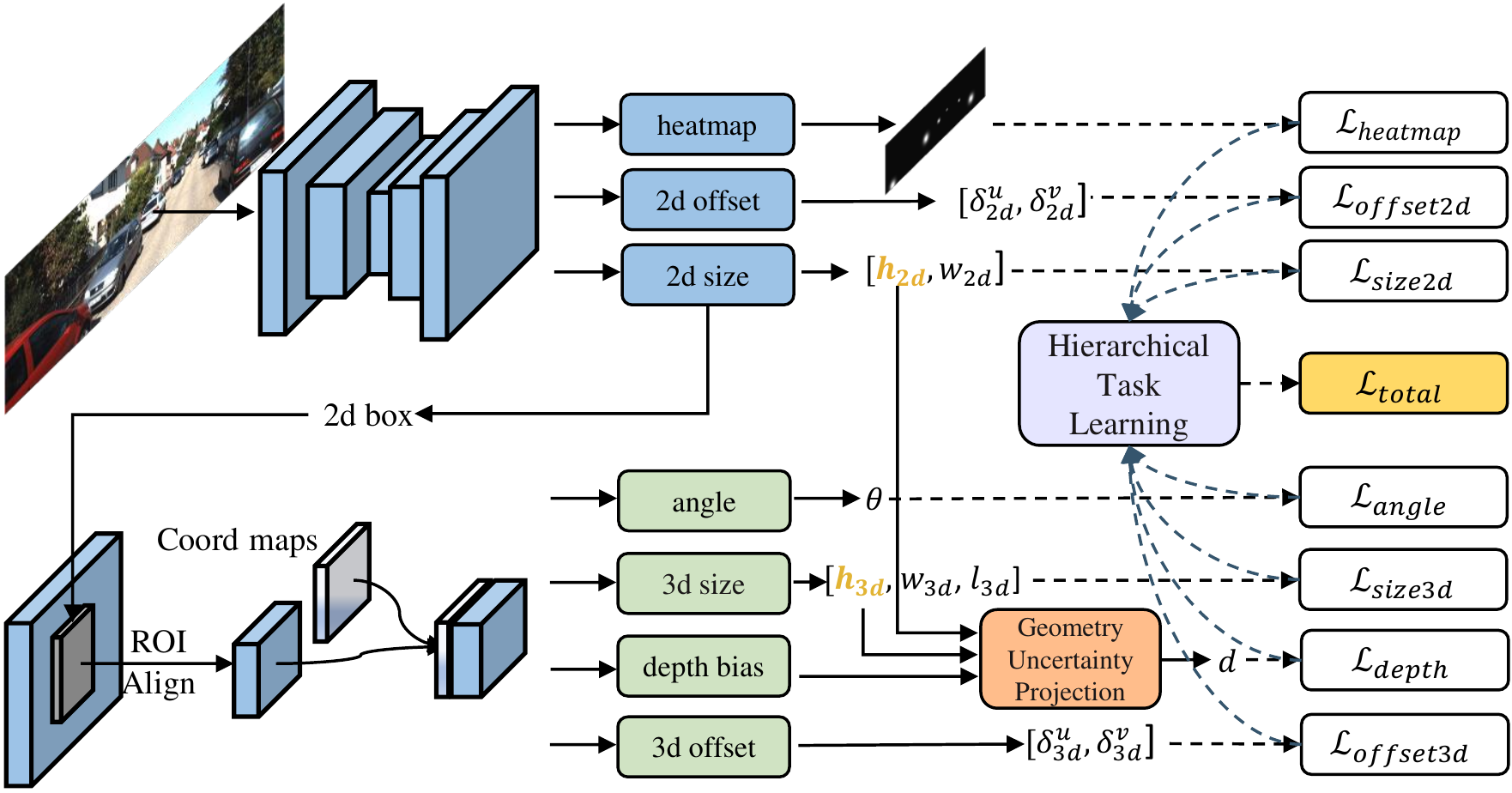}
\end{center}
   \caption{The framework of the Geometry Uncertainty Projection Network. The input image is sent to the network to extract the 2D box and basic 3D box parameters. And the Geometry Uncertainty Projection module would infer the depth based on the height parameters. And at the training, all task losses would be gathered by the Hierarchical Task Learning strategy to assign proper weights for them. }
\label{fig:pipeline}
\end{figure*}

\noindent 
\textbf{Monocular 3D object detection.} The monocular 3D object detection aims to predict 3D bounding boxes from a single given image~\cite{ding2020learning,he2019mono3d++,kundu20183d,liu2019deep,manhardt2019roi,simonelli2019disentangling}. 
Existing methods focus on deep representation learning~\cite{roddick2018orthographic} and geometry priors~\cite{ma2020rethinking,ma2019accurate,wang2019pseudo,ma2021delving}. Deep3DBox~\cite{mousavian20173d} firstly tried to solve the key angle prediction problem by geometry priors.  DeepMANTA~\cite{chabot2017deep} introduced the 3D CAD model to learn shape-based knowledge and guided to better dimension prediction results.  GS3D~\cite{li2019gs3d} utilized the ROI surface features to extract better object representations.  M3DRPN~\cite{brazil2019m3d} gave a novel modified 3D anchor setting and proposed a depth-wise convolution to treat the monocular 3D detection task.  MonoPair~\cite{chen2020monopair} proposed a pair-wise relationship to improve the monocular 3D detection performance.

Except for these methods, many methods tried to introduce geometry projection to infer depth~\cite{bao2020object,barabanau2019monocular,cai2020monocular,ku2019monocular}. Ivan~\emph{et al.}~\cite{barabanau2019monocular} combined the keypoint method and the projection to do geometry reasoning.  
Decoupled3D~\cite{cai2020monocular} used lengths of bounding box edges to project and get the inferred depth. Bao~\emph{et al.}~\cite{bao2020object} combined the center voting with the geometry projection to achieve better 3D center reasoning. All of these projection-based mono3D methods did not consider the error amplification problem, leading to the limited performance. 

\noindent 
\textbf{Uncertainty based depth estimation.} The uncertainty theory is widely used in the deep regression method~\cite{blundell2015weight}, which can model both aleatoric and epistemic uncertainty. This technology is well developed in the depth estimation~\cite{kendall2017uncertainties,liu2019neural}, which can significantly reduce the noise of the depth targets. However, these methods directly regressed the depth uncertainty by the deep models and neglected the relationships between the height and the depth. In this work, we try to compute the uncertainty via combining both end-to-end learning and the geometry relationships.


\noindent \textbf{Multi-task learning.} Multi-task learning is a widely studied topic in computer vision. Many works focus on task relation representation learning~\cite{long2015learning,vandenhende2020mti,xu2018pad,zamir2020robust,zamir2018taskonomy,zhang2019pattern}. Except that, some works also tried to adjust weights for different loss functions to solve the multi-task problem~\cite{chen2018gradnorm,kendall2018multi,zhang2014facial}. 
GradNorm~\cite{chen2018gradnorm} tried to solve the loss unbalance problem in joint multi-task learning and improved the training stability. Kendall~\emph{et al.}~\cite{kendall2018multi} proposed a task-uncertainty strategy to treat the task balance problems, which also achieved good results. These loss weights control methods assumed that each task is independent from each other, which are unsuitable for our method, since the multiple tasks in our framework form a hierarchical structure, i.e., some tasks are dependent on their pre-tasks. Therefore, we propose a Hierarchical Task Learning strategy to handle it in this work.

\section{Geometry Uncertainty Projection Network}
Figure~\ref{fig:pipeline} shows the framework of the proposed Geometry Uncertainty Projection Network (GUP Net). It takes an image as input and processes it with a 2D detection backbone first, yielding 2D bounding boxes (Region of Interest, RoI) and then computes some basic 3D bounding box information,\textit{i.e.}, angle, dimensions and 3D projected center for each box. After that, the Geometry Uncertainty Projection (GUP) module predicts the depth distribution via combining both mathematical priors and uncertainty modeling. This depth distribution provides an accurate inferred depth value and its corresponding uncertainty. The predicted uncertainty would be mapped to 3D detection confidence at the inference stage. Furthermore, to avoid misleading caused by the error amplification at the beginning of training, an efficient Hierarchical Task Learning (HTL) strategy would control the overall training process, where each task does not start training until its pre-tasks have been trained well. 

\subsection{2D detection}
Our 2D detector is built on CenterNet~\cite{zhou2019objects}, which includes a backbone network and three 2D detection sub-heads to compute the location, size, and confidence for each potential 2D box.
As shown in Figure~\ref{fig:pipeline}, the heatmap head computes a heatmap with the size of $(W\times H\times C)$ to indicate the coarse locations and confidences of the objects in the given image, where $C$ is the number of categories.
Based on that, a 2D offset branch computes the bias $(\delta_{2d}^u,\delta_{2d}^v)$ to refine the coarse locations to the accurate bounding box center, and a 2D size branch predicts the size $(w_{2d},h_{2d})$ for each box. 
Their loss functions are denoted as $\mathcal{L}_{heatmap}$, $\mathcal{L}_{offset2d}$ and $\mathcal{L}_{size2d}$. 


\subsection{RoI feature representation}
To guide the model to focus on the object, we crop and resize the RoI features using RoIAlign~\cite{he2017mask}.
The RoI features only contain the object-level features and do not include the background noise. 
However, those features lack location and size cues which are essential to the monocular depth estimation~\cite{dijk2019neural}. 
Therefore, we compute the normalized coordinate map, and then concatenate it with the feature maps of each RoI in a channel-wise manner to compensate for that cues (shown as Figure~\ref{fig:pipeline}).

\subsection{Basic 3D detection Heads}
With the extracted RoI features, we construct several sub-heads on top of those features to predict some basic 3D bounding box information. A 3D offset branch aims to estimate the 3D center projection on the 2D feature maps~\cite{chen2020monopair}. The angle prediction branch predicts the relative alpha rotation angle~\cite{mousavian20173d}. And the 3D size branch estimates the 3D dimension parameters, including height, width and length. These predictions are supervised by $\mathcal{L}_{offset3d}$, $\mathcal{L}_{angle}$ and $\mathcal{L}_{size3d}$, respectively. Note that $\mathcal{L}_{size3d}$ includes three parts for different dimensions, \textit{e.g.}, the height loss $\mathcal{L}_{h3d}$. 

\subsection{Geometry Uncertainty Projection}
The basic 3D detection heads provide most information of the 3D bounding box except depth. Given the difficulty to regress depth directly, we propose a novel Geometry Uncertainty Projection model. The overall module builds the projection process in the probability framework rather than single values so that the model can compute the theoretical uncertainty for the inferred depth, which can indicate the depth inference reliability and also be helpful for the depth learning. 

To achieve this goal, we first assume the prediction of the 3D height for each object is a Laplace distribution $La(\mu_{h},\lambda_{h})$ \footnote{The probability density function of a Laplace random variable $X\sim La(\mu,\lambda)$ is: $ f_X(x)= \frac{1}{2\lambda}\text{exp}(\frac{|x-\mu|}{\lambda})$,
where $\mu$ and $\lambda$ are parameters of the Laplace distribution. The standard deviation $\sigma$ has a relationship with the parameters $\lambda$: $\sigma = \sqrt{2}\lambda$. 
}. The distribution parameters $\mu_{h}$ and $\sigma_{h}$ are predicted by the 3D size stream in an end-to-end way. The $\mu_{h}$ denotes the regression target output and the $\sigma_{h}$ is the uncertainty of the inference. Consequently, the 3D height loss function can be defined as:
\begin{equation}
\begin{aligned}
    \mathcal{L}_{h3d} =
    \frac{\sqrt{2}}{\sigma_{h}}|\mu_{h}-h^{gt}_{3d}|+log(\sigma_{h}).
\end{aligned}
\end{equation}
The minimization of $\mathcal{L}_{h3d}$ make $\mu_{h}$ and the ground-truth height $h^{gt}_{3d}$ as close as possible. Particularly, the difficult or noise-labeled samples usually incur large $\sigma_{3d}$, indicating the low prediction confidence. Based on the learned $h_{3d}$ distribution, the depth distribution of the projection output $La(\mu_{p},\lambda_{p})$ can be approximated as:
\begin{equation}
\begin{aligned}
    d_{p} &= \frac{f\cdot h_{3d}}{h_{2d}}
    = \frac{f\cdot (\lambda_{h}\cdot X + \mu_{h})}{h_{2d}}\\
    & = \frac{f\cdot \lambda_{h}}{h_{2d}}\cdot X + \frac{f\cdot \mu_{h}}{h_{2d}},
\end{aligned}
\end{equation}
where $X$ is the standard Laplace distribution $La(0,1)$. In this sense, the mean and standard deviation of the projection depth $d_p$ are $\frac{f\cdot \mu_{h}}{h_{2d}}$ and $\frac{f\cdot \sigma_{h}}{h_{2d}}$, respectively. 
To obtain better predicted depth, we add a learned bias to modify the initial projection results. We also assume that the learned bias is a Laplace distribution $La(\mu_{b},\lambda_{b})$ and independent with the projection one. Accordingly, the final depth distribution can be written as:
\begin{equation}
\begin{aligned}
     d = La(\mu_{p},\sigma_{p}) &+ La(\mu_{b},\sigma_{b}),\\
     \mu_d = \mu_{p}+\mu_{b},\quad \sigma_{d} &= \sqrt{(\sigma_{p})^2+(\sigma_{b})^2}.
\end{aligned}
\end{equation}
We refer to the final uncertainty $\sigma_{d}$ as Geometry based Uncertainty (GeU). This uncertainty reflects both the projection uncertainty and the bias learning uncertainty. With this formula, a small uncertainty of $h_{3d}$ will be reflected in the GeU value. To optimize the final depth distribution, we apply the uncertainty regression loss:
\begin{equation}
\label{eq:depth_loss}
\begin{aligned}
    \mathcal{L}_{depth} = \frac{\sqrt{2}}{\sigma_{d}}|\mu_{d}-d^{gt}|+log(\sigma_{d}).
\end{aligned}
\end{equation}
Note that we also assume the depth distribution belong to Laplace distribution here for simplification. The overall loss would push the projection results close to the ground truth $d^{gt}$ and the gradient would affect the depth bias, the 2D height and the 3D height simultaneously. Besides, the uncertainty of 3D height and depth bias is also trained in the optimization process.

During inference, the reliability of depth prediction is critical for real-world applications. A reliable inference system is expected to feedback high confidence for a good estimation and low score for a bad one. As our well-designed GeU has capability of indicating the uncertainty of depth, we further map it to a value between 0$\sim$1 by an exponential function to indicate the depth Uncertainty-Confidence (UnC):
\begin{equation}
\label{eq:unc}
\begin{aligned}
     p_{depth} = exp(-\sigma_{d}).
\end{aligned}
\end{equation}
It can provide more accurate confidence for each projection depth. Thus we use this confidence as the conditional 3D bounding box scores $p_{3d|2d}$ in the testing. The final inference score can be computed as:
\begin{equation}
\begin{aligned}
    p_{3d} = p_{3d|2d}\cdot p_{2d} = p_{depth}\cdot p_{2d}.
\end{aligned}    
\end{equation}
This score represents both the 2D detection confidence and the depth inference confidence, which can guide better reliability. 


\subsection{Hierarchical Task Learning}
The GUP module mainly addresses the error amplification effect in the inference stage. Yet, this effect also damages the training procedure. Specifically, at the beginning of the training, the prediction of both $h_{2d}$ and $h_{3d}$ are far from accurate, which will mislead the overall training and damage the performance.  
To tackle this problem, we design a Hierarchical Task Learning (HTL) to control weights for each task at each epoch. The overall loss is:
\begin{equation}
\begin{aligned}
     \mathcal{L}_{total} = \sum_{i\in \mathcal{T}} w_{i}(t)\cdot \mathcal{L}_{i},
\end{aligned}
\end{equation}
where $\mathcal{T}$ is the task set. $t$ denotes the current epoch index and $\mathcal{L}_{i}$ means the $i$-th task loss function. $w_{i}(t)$ is the loss weight for the $i$-th task at the $t$-th epoch.

\begin{figure}[t]
\begin{center}
\includegraphics[width=1.0\linewidth]{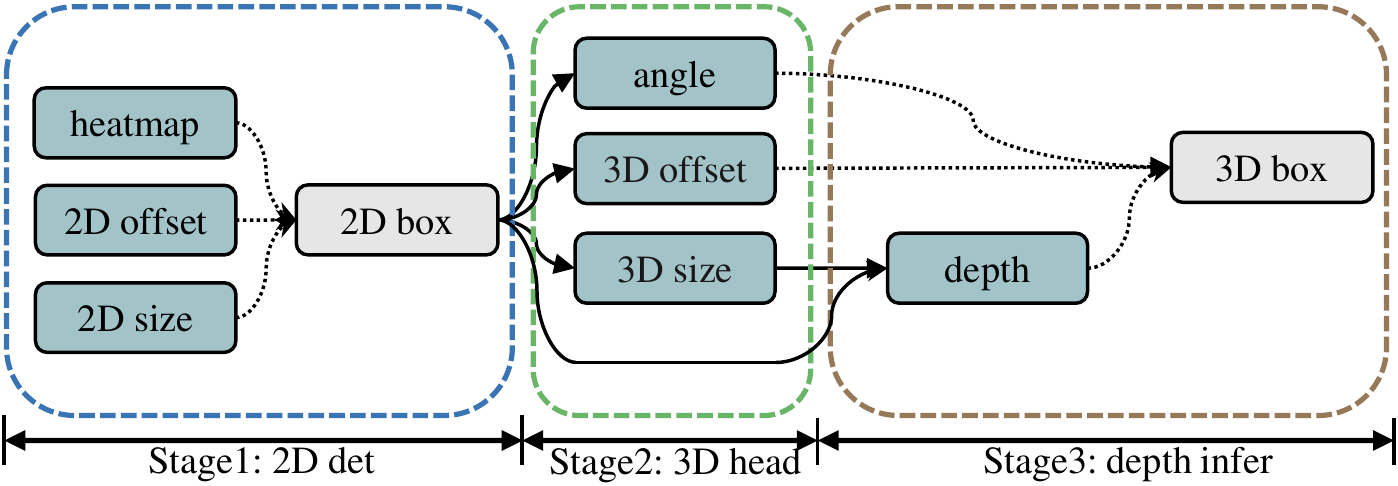}
\end{center}
   \caption{The task hierarchy of the GUP Net. The first stage is
2D detection. Built on top of RoI features, the second stage
consists of basic 3D detection heads. Based on 2D and 3D heights
estimated in the previous stages, the third stage infers the depth and
then constitutes the 3D bounding box.}
\label{fig:task_graph}
\end{figure}

\begin{figure}[t]
\begin{center}
\includegraphics[width=0.8\linewidth]{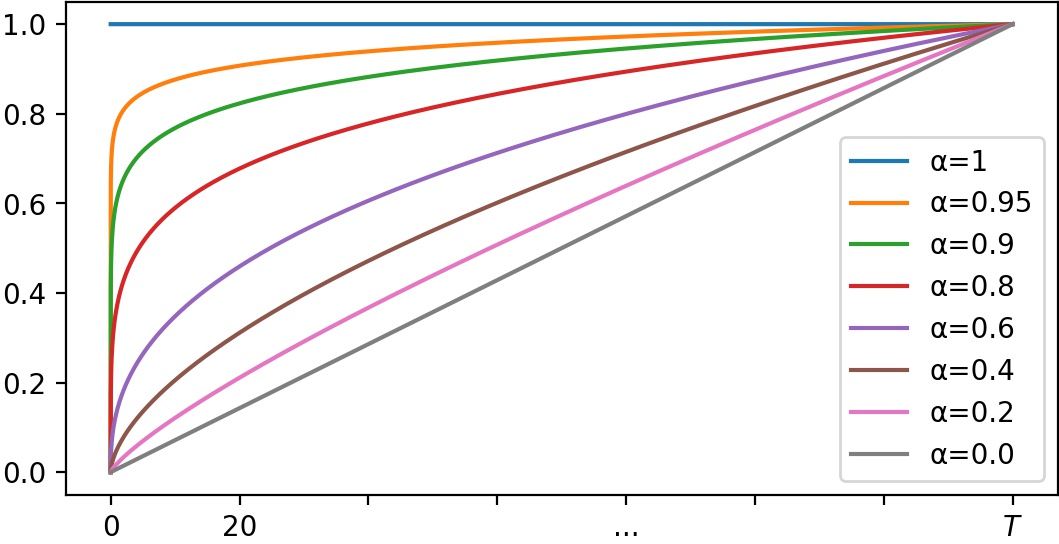}
\end{center}
   \caption{The polynomial time scheduling function with the adjust parameter. The vertical axis is the value of $w_i(t)$ and the horizontal axis is the epoch index $t$. (best viewed in color.)}
\label{fig:time_schedule}
\end{figure}

HTL is inspired by the motivation that each task should start training after its pre-task has been trained well. We split tasks into different stages as shown in Figure~\ref{fig:task_graph} and the loss weight $w_{i}(t)$ should be associated with all pre-tasks of the $i$-th task. The first stage is 2D detection, including heatmap, 2D offset, 2D size. Then, the second stage is the 3D heads containing angle, 3D offset and 3D size. All of these 3D tasks are built on the ROI features, so the tasks in 2D detection stage are their pre-tasks. Similarly, the final stage is the depth inference and its pre-tasks are the 3D size and all the tasks in 2D detection stage since depth prediction depends on the 3D height and 2D height.
To train each task sufficiently, we aim to gradually increase the $w_{i}(t)$ from 0 to 1 as the training progresses. So we adopt the widely used polynomial time scheduling function~\cite{morerio2017curriculum} in the curriculum learning topic as our weighted function, which is adapted as follows:
\begin{equation}
\begin{aligned}
     w_{i}(t) = (\frac{t}{T})^{1-{\alpha_i(t)}},\ \ \alpha_i(t) \in [0,1],
\end{aligned}
\end{equation}
where $T$ is the total training epochs and the normalized time variable $\frac{t}{T}$ can automatically adjust the time scale. ${\alpha_i}(t)$ is an adjust parameter at the $t$-th epoch, corresponding to every pre-task of the $i$-th task. Figure~\ref{fig:time_schedule} shows that ${\alpha_i}$ can change the trend of the time scheduler. The larger ${\alpha_i}$ is, the faster $w_{i}(\cdot)$ increases.
\begin{table*}[!ht]
\centering
\fontsize{8}{10}\selectfont
\caption{{\bf 3D object detection on the KITTI \emph{test} set.} We highlight the best results in {\bf bold}. For the extra data: 1). `Depth' means the methods use extra depth annotations or off-the-shelf networks pre-trained from a larger depth estimation dataset. 2). `Temporal' means using additional temporal  data. 3). `LiDAR' means utilizing real LiDAR data for better training. 4). `None' denotes no extra data is used.}
\label{tab:kitti_test}
\begin{tabular}{l||c|ccc|ccc|ccc}
\toprule
\multirow{2}{*}{Method}&\multirow{2}{*}{Extra data}&
\multicolumn{3}{c|}{Car@IoU=0.7}&\multicolumn{3}{c|}{Pedestrian@IoU=0.5}&
\multicolumn{3}{c}{Cyclist@IoU=0.5} \cr\cline{3-11} & &
Easy&Mod.&Hard&Easy&Mod.&Hard&Easy&Mod.&Hard\cr\hline
Mono-PLiDAR~\cite{weng2019monocular}&Depth
& 10.76 & 7.50 & 6.10
& – & – &–
& – & – &–  \cr
Decoupled-3D~\cite{cai2020monocular}&Depth
& 11.08 & 7.02 & 5.63
& – & – & –
& – & – & –\cr
AM3D~\cite{ma2019accurate}&Depth
& 16.50 & 10.74 & 9.52 
& – & – & – 
& – & – & –\cr
PatchNet~\cite{ma2020rethinking}&Depth
& 15.68 & 11.12 & 10.17
& – & – & –
& – & – & –  \cr
DA-3Ddet~\cite{da3dnet}&Depth
& 16.77 & 11.50 & 8.93
& – & – & –
& – & – & –\cr
D4LCN~\cite{ding2020learning}&Depth
& 16.65 & 11.72 & 9.51
& 4.55 & 3.42 & 2.83
& 2.45 & 1.67 & 1.36\cr
Kinematic~\cite{brazil2020kinematic}&Temporal
& 19.07 & 12.72 & 9.17
& – & – & –
& – & – & –\cr
MonoPSR~\cite{ku2019monocular}&LiDAR
& 10.76 & 7.25 & 5.85
& 6.12 & 4.00 & 3.30
& \textbf{8.70} & \textbf{4.74} & \textbf{3.68}\cr
CaDNN~\cite{reading2021categorical}&LiDAR
& 19.17 & 13.41 & 11.46
& 12.87 & 8.14 & 6.76
& 7.00 & 3.41 & 3.30 \cr\hline
MonoDIS~\cite{simonelli2019disentangling}&None
& 10.37 & 7.94 & 6.40
& – & – & –
& – & – & –\cr
UR3D~\cite{ur3d}&None
& 15.58 & 8.61 & 6.00
& – & – & –
& – & – & –\cr
M3D-RPN~\cite{brazil2019m3d}&None
& 14.76 & 9.71 & 7.42
& 4.92 & 3.48 & 2.94
& 0.94 & 0.65 & 0.47\cr
SMOKE~\cite{smoke}&None
& 14.03 & 9.76 & 7.84
& – & – & –
& – & – & –\cr
MonoPair~\cite{chen2020monopair}&None
& 13.04 & 9.99 & 8.65
& 10.02 & 6.68 & 5.53
& 3.79 & 2.12 & 1.83\cr
RTM3D~\cite{li2020rtm3d}&None
& 14.41 & 10.34 & 8.77
& – & – & –
& – & – & –\cr
MoVi-3D~\cite{movi3d}&None
& 15.19 & 10.90 & 9.26
& 8.99 & 5.44 & 4.57
& 1.08 & 0.63 & 0.70\cr
RAR-Net~\cite{rarnet}&None
& 16.37 & 11.01 & 9.52
& – & – & –
& – & – & –\cr 
\hline 
GUP Net (Ours)&None
&\textbf{20.11}&\textbf{14.20}&\textbf{11.77}
&\textbf{14.72}&\textbf{9.53}&\textbf{7.87}
& 4.18 & 2.65 & 2.09 \cr
Improvement&\vs Depth
&+3.46 & +2.48 & +2.26
&+10.17 & +6.11 & +5.04
&+1.73 & +0.98 & +0.73\cr
Improvement& \vs Temporal
&+1.04 & +1.48 & +2.60
& – & – & –
& – & – & –\cr
Improvement&\vs LiDAR
& +0.94 & +0.79 & +0.31
& +1.85 & +1.39 & +1.11
& -4.52 & -2.09 & -2.09\cr
Improvement&\vs None
& +3.74 & +3.19 & +2.25
& +4.7 & +2.85 & +2.34
& +0.39 & +0.53 & +0.26\cr
\bottomrule  
\end{tabular}
\end{table*}
From the definition of the adjust parameter, it is natural to decide its value via the learning situation of every pre-task. If all pre-task have been well trained, the ${\alpha_i}$ is expected be large, otherwise it should be small. This is motivated by the observation that human usually learn advanced courses after finishing fundamental courses. Therefore, ${\alpha_i}(t)$ is defined as:
\begin{equation}
\begin{aligned}
     {\alpha_i}(t) = \prod_{j\in{\mathcal{P}_i}}ls_j(t),
\end{aligned}
\end{equation}
where $\mathcal{P}_i$ is the pre-task set for the $i$-th task. $ls_j$ means the learning situation indicator of $j$-th task, which is a value between 0$\sim$1. This formula means that $\alpha_i$ would get high values only when all pre-tasks have achieved high $ls$ (trained well). For the $ls_j$, inspired by~\cite{chen2018gradnorm,zhang2014facial}, we design a scale-invariant factor to indicate the learning situation:
\begin{equation}
\begin{aligned}
     {ls_j}(t) &= \frac{\mathcal{DF}_j(K)-\mathcal{DF}_j(t)}{\mathcal{DF}_j(K)},\\
     \mathcal{DF}_j(t) &= \frac{1}{K}\mathop{\sum}\limits_{\hat{t}=t-K}^{t-1}|\mathcal{L}'_j\left( \hat{t}\right)|,
\end{aligned}
\end{equation}
where $\mathcal{L}'_j(\hat{t})$ is the derivative of the $\mathcal{L}_j(\cdot)$ at the $\hat{t}$-th epoch, which can indicate the local change trend of the loss function. The $\mathcal{DF}_j(t)$ computes the mean of derivatives in the recent $K$ epochs before the $t$-th epoch to reflect the mean change trend. If the $\mathcal{L}_j$ drops quickly in the recent $K$ epochs, the $\mathcal{DF}_j$ will get a large value. So the ${ls_j}$ formula means comparing the difference between the current trend $\mathcal{DF}_j(t)$ and the trend of the first $K$ epochs at the beginning of training $\mathcal{DF}_j(K)$ for the $j$-th task. If the current loss trend is similar to the beginning trend, the indicator will give a small value, which means that this task has not trained well. Conversely, if a task tends to converge, the ${ls_j}$ will be close to 1, meaning that the learning situation of this task is satisfied.


Based on the overall design, the loss weight of each term can reflect the learning situation of its pre-tasks dynamically, which can make the training more stable.


\section{Experiments}
\subsection{Setup}
\noindent
\textbf{Dataset.} The KITTI 3D dataset~\cite{kitti} is the most commonly used benchmark in the 3D object detection task, and it provides left camera images,  calibration files, annotations for standard monocular 3D detection. 
It totally provides 7,481 frames for training and 7,518 frames for testing.
Following~\cite{mono3d,mv3d}, we split the training data into a training set (3,712 images) and a validation set (3,769 images). 
We conduct ablation studies based on this split and also report the final results with the model trained on all 7,481 images and tested by KITTI official server.

\noindent
\textbf{Evaluation protocols.} All the experiments follow the standard evaluation protocol in the monocular 3D object detection and bird's view (BEV) detection tasks. 
Following~\cite{simonelli2019disentangling}, we evaluate the ${\rm AP}_{40}$ to avoid the bias of original ${\rm AP}_{11}$. 

\noindent
\textbf{Implementation details.} We use DLA-34~\cite{yu2018deep} as our backbone for both baseline and our method. The resolution of the input image is set to 380 × 1280 and the feature maps down-sampling rate is 4. Each 2D sub-head has two Conv layers (the channel of the first one is set to 256) and each 3D sub-head includes one 3x3 Conv layer with 256 channels, one averaged pooling layer and one fully-connected layer. The output channels of these heads are depending on the output data structure. We train our model with the batchsize of 32 on 3 Nvidia TiTan XP GPUs for 140 epochs. The initial learning rate is 1.25$e^{-3}$, which is decayed by 0.1 at the 90-th and the 120-th epoch. To make the training more stable, we apply the linear warm-up strategy in the first 5 epochs. The $K$ in the HTL is also set to 5.

\subsection{Main Results}

\begin{table*}[!t]
\begin{center}
\fontsize{8}{10}\selectfont
\caption{{\bf Performance of the Car category on the KITTI \emph{validation} set.} 
We highlight the best results in {\bf bold}.}
\label{tab:kitti_val}
\begin{tabular}{l||ccc|ccc|ccc|ccc}
\toprule
\multirow{2}{*}{Method} & \multicolumn{3}{c|}{3D@IoU=0.7} & \multicolumn{3}{c|}{BEV@IoU=0.7} & \multicolumn{3}{c|}{3D@IoU=0.5} & \multicolumn{3}{c}{BEV@IoU=0.5}\\ 
\cline{2-13} 
 ~ & Easy & Mod. & Hard  & Easy & Mod. & Hard & Easy & Mod. & Hard & Easy & Mod. & Hard \\ 
\hline
CenterNet~\cite{zhou2019objects} 
& 0.60 & 0.66 & 0.77 
& 3.46 & 3.31 & 3.21 
& 20.00 & 17.50 & 15.57
& 34.36 & 27.91 & 24.65 \\  
MonoGRNet~\cite{qin2019monogrnet} 
& 11.90 & 7.56  & 5.76 
& 19.72 & 12.81 & 10.15 
& 47.59 & 32.28 & 25.50
& 48.53 & 35.94 & 28.59 \\  
MonoDIS~\cite{simonelli2019disentangling}  
& 11.06 & 7.60 & 6.37 
& 18.45 & 12.58 & 10.66 
& - & - &
& - & - &\\  
M3D-RPN~\cite{brazil2019m3d}
& 14.53 & 11.07 & 8.65 
& 20.85 & 15.62 & 11.88 
& 48.53 & 35.94 & 28.59
& 53.35 & 39.60 & 31.76\\ 
MoVi-3D~\cite{movi3d}	
&14.28  &11.13  &9.68      
&22.36  &17.87  &15.73
&-		&-		&-		
&-		&-		&-	\\
MonoPair~\cite{chen2020monopair}
& 16.28 & 12.30 & 10.42
& 24.12 & 18.17 & 15.76
& 55.38 & {\bf 42.39} & {\bf 37.99}
& 61.06 & {\bf 47.63} & {\bf 41.92}\\ \hline
GUP Net (Ours)   
& {\bf 22.76} & {\bf 16.46} & {\bf 13.72} 
& {\bf 31.07} & {\bf 22.94} & {\bf 19.75} 
& {\bf 57.62} & 42.33 & 37.59 
& {\bf 61.78} & 47.06 & 40.88\cr 
\bottomrule
\end{tabular}
\end{center}
\end{table*}

\begin{table}[!t]
    \centering
    \fontsize{7}{10}\selectfont
    \caption{{\bf Ablation studies} on the KITTI \emph{validation} set for the Car category.}
    \label{tab:ablation}
    \renewcommand\tabcolsep{3.0pt}
    \resizebox{\linewidth}{!}{
    \begin{tabular}{l|ccccc|ccc|ccc}
        \toprule
        \multirow{2}{*}{}&\multirow{2}{*}{CM}&\multirow{2}{*}{UnC}&\multirow{2}{*}{GeP}&\multirow{2}{*}{GeU}&\multirow{2}{*}{HTL}&
        \multicolumn{3}{c|}{3D@IoU=0.7}&\multicolumn{3}{c}{BEV@ IoU=0.7}\cr\cline{7-12} &
         & & & & &Easy&Mod.&Hard&Easy&Mod.&Hard\cr\hline
        (a)&-&-&-&-&-&15.18&11.00&9.52&21.57&16.43&13.93\cr
        (b)&\checkmark&-&-&-&-&16.39&12.44&11.01&23.08&18.32&16.03\cr
        (c)&\checkmark&\checkmark&-&-&-&19.69&13.53&11.33&27.49&19.00& 16.96\cr 
        (d)&\checkmark&-&\checkmark&-&-&17.27&12.79&10.51&24.02&18.73&15.07\cr
        (e)&\checkmark&\checkmark&\checkmark&-&-&18.23&13.57&11.22&26.17& 19.19&16.15\cr
        (f)&\checkmark&\checkmark&\checkmark&\checkmark&-&20.86&15.70& 13.21&27.54&20.80&17.77\cr
        (g)&\checkmark&\checkmark&\checkmark&-&\checkmark&21.00&15.63& 12.98&30.03&21.32&18.17\cr 
        (h)&\checkmark&\checkmark&\checkmark&\checkmark&\checkmark&\textbf{22.76}&\textbf{16.46}&\textbf{13.72}&\textbf{31.07}&\textbf{22.94}&\textbf{19.75}\cr
        \bottomrule        
    \end{tabular}}
\end{table}

\noindent
{\bf Results of Car category on the KITTI \emph{test} set.}
As shown in Table~\ref{tab:kitti_test}, we first compare our method with other counterparts on the KITTI test set.
Overall,  the proposed method achieves superior results of Car category over previous methods, including those with extra data.
Under fair conditions, our method achieves 3.74\%, 3.19\%, and 2.25\% gains on the easy, moderate, and hard settings, respectively.
Furthermore, our method also outperforms the methods with extra data.
For instance, compared with the recently proposed CaDNN~\cite{reading2021categorical} utilizing LiDAR signals as supervision of depth estimation sub-task, our method still obtains 0.94\%, 0.79\%, and 0.31\% gains on the three difficulty settings, which confirms the effectiveness of the proposed method.

\noindent
{\bf Results of Car category on the KITTI \emph{validation} set.}
We also present our model’s performance on the KITTI validation set in Table~\ref{tab:kitti_val} for better comparison, including different tasks and IoU thresholds.
Specifically, our method gets almost the same performance as the best competing method MonoPair at the 0.5 IoU threshold. 
Moreover, our method improves with respect to MonoPair by 4.16\%/4.77\% for 3D/BEV detection under the moderate setting at 0.7 IoU threshold.
This shows that our method is very suitable for high-precision tasks, which is a vital feature in the automatic driving scene.
Note that RTM3D and RAR-Net do not report the ${\rm AP}_{40}$ metric on the validation set, and the comparison with them on ${\rm AP}_{11}$ metric can be found in supplementary materials.

\noindent
{\bf Pedestrian/Cyclist detection on the KITTI \emph{test} set.}
We also report the Pedestrian/Cyclist detection results in Table~\ref{tab:kitti_test}. 
Specifically, our method remarkably outperforms all the competing methods on all levels of difficulty for pedestrian detection. 
As for cyclist detection, our approach is superior to other methods except for MonoPSR and CaDNN.
The main reason is those two methods can benefit from the extra depth supervision derived from LiDAR signals, thereby improving the overall performance.
In contrast, the performances of the others are limited by the few training samples (there are 14,357/2,207/734 instances in total in the KITTI train-val set).
It should be noted that our method still ranks first for methods without extra data.

\noindent
{\bf Latency analysis.}
We also test the running time of our system. We test the averaged running time on a single Nvidia TiTan XP GPU and achieve 29.4 FPS, which shows the efficiency of the inference pipeline.

\subsection{Ablation Study} 
To understand how much improvement each component provides, we perform ablation studies on the KITTI validation set for the Car category, and the main results are summarized in Table~\ref{tab:ablation}. 

\noindent
{\bf Effectiveness of the Coordinate Maps}.
We concatenate a Coordinate Map (CM) for each RoI feature, and the experiment (a$\rightarrow$b) clearly shows the effectiveness of this design, which means the location and size cues are crucial to our task.
Note that the additional computing overhead introduced by CM is negligible.
\begin{figure*}[t]
\begin{center}
\includegraphics[width=1.0\linewidth]{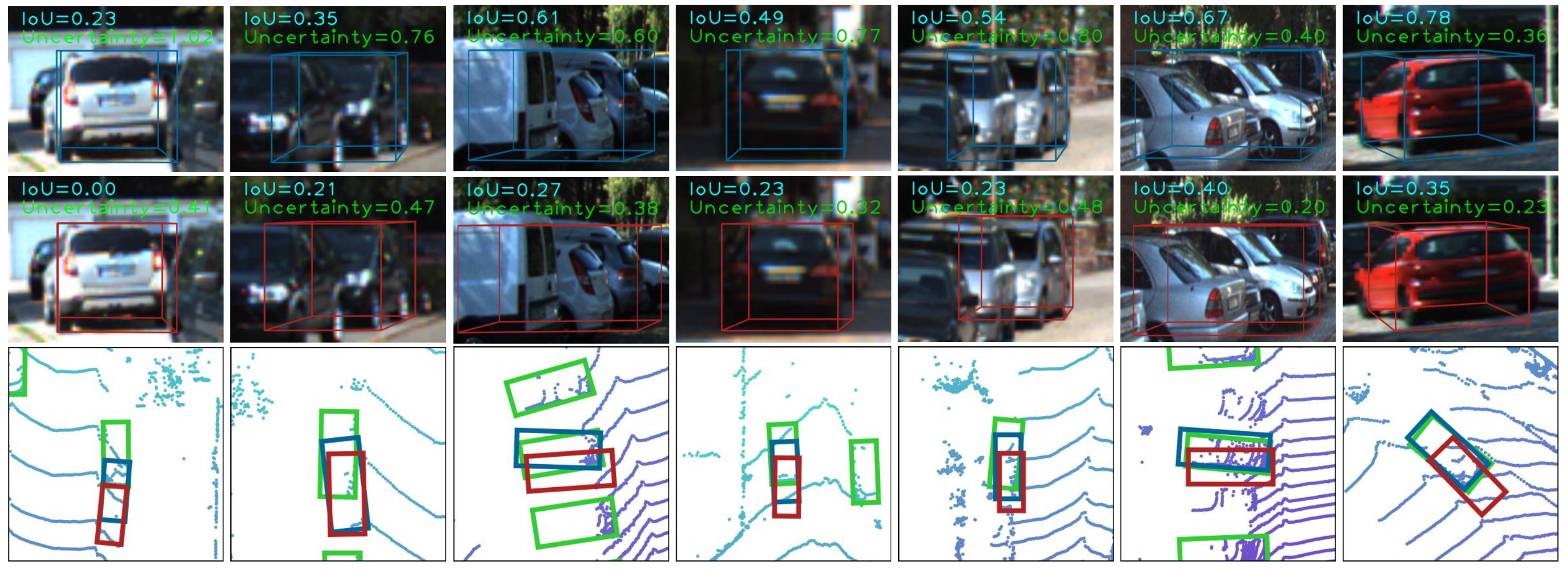}
\end{center}
   \caption{The visualized uncertainty examples on the validation set. The first row (\textcolor[RGB]{0,102,153}{Blue} boxes) are results of our method. The second row (\textcolor[RGB]{178,34,34}{Red} boxes) is the baseline results. The 3rd row shows the bird-view results (\textcolor[RGB]{50,205,50}{Green} means the ground truth boxes). The IoU means the Intersection-over-Union between the predicted box and the corresponding ground-truth one. The uncertainty value is equal to the standard deviation (best viewed in color.).}
\label{fig:uncertain}
\end{figure*}
\begin{figure}[t]
\begin{center}
\includegraphics[width=1.0\linewidth]{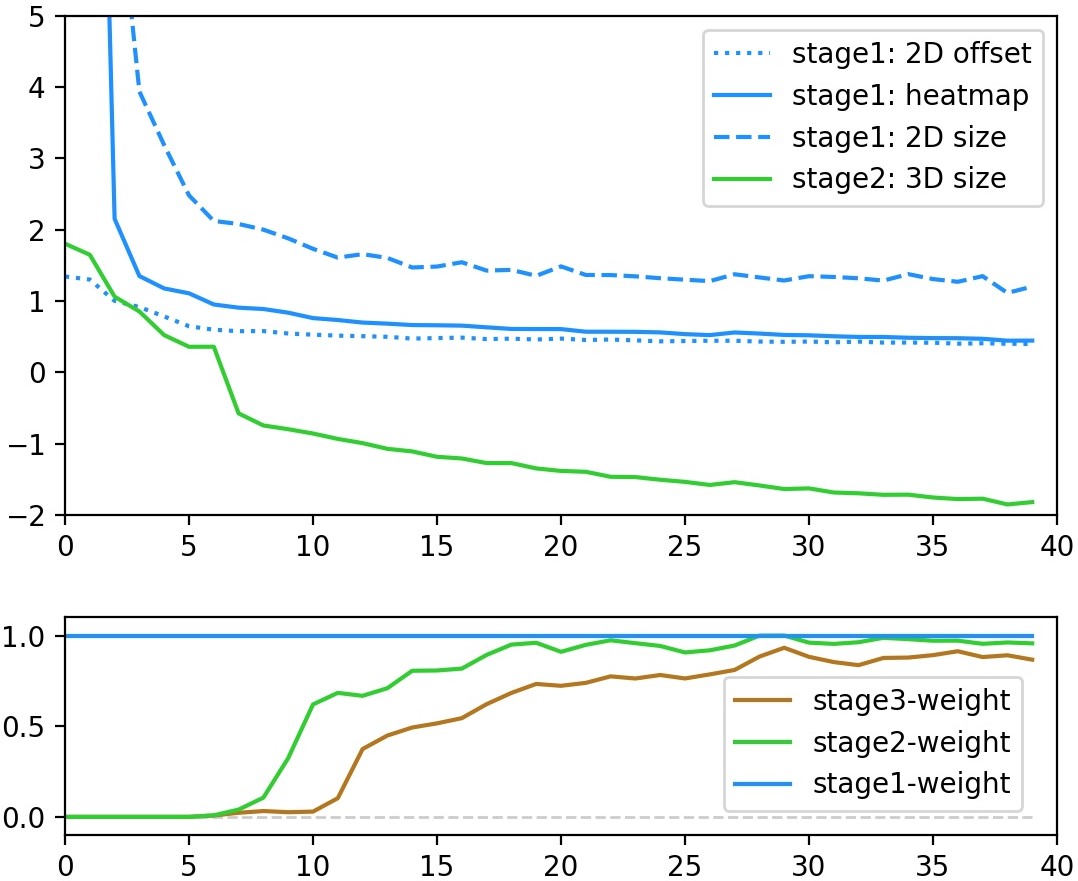}
\end{center}
   \caption{The upper image shows the loss curves and the bottom image means the loss weight trends. The blue, green and brown mean the 1st, 2nd and 3rd stages in the Figure~\ref{fig:task_graph} respectively.}
\label{fig:loss_weight}
\end{figure}

\noindent
{\bf Comparison of the Geometry Uncertainty Projection}.
We evaluate our Geometry Uncertainty Projection (GUP) module here. Note that, we think our GUP module brings gains from the following parts: geometry projection (GeP), Geometry based Uncertainty (GeU) and the Uncertainty-Confidence (UnC, Eq.~\ref{eq:unc}). So we evaluate the effectiveness of these three parts respectively. 
First, we evaluate the effectiveness of the UnC. By comparing settings (b$\rightarrow$c and d$\rightarrow$e), we can find the UnC part can effectively and stably improve the overall performance, \eg 1.09\% improvement for (b$\rightarrow$c) and 0.78\% improvement for (d$\rightarrow$e) on 3D detection task under moderate level. After that, we concern the GeP part effectiveness, we can see that adding GeP part improves the performance in the experiment (b$\rightarrow$d) without UnC, but leads to an accuracy drop in the experiment (c$\rightarrow$e) with UnC (The c and e experiments use directly learning uncertainty in the Eq.~\ref{eq:unc} to indicate confidence).
This proves our motivation. It is hard for the projection-based model to directly learn accurate uncertainty and confidence because of the error amplification. 
Furthermore, note that the accuracy of hard cases decreases in both groups of experiments, which indicates the traditional projection cannot deal with the difficult cases caused by heavy occlusion/truncation.
Second, we apply our GeU strategy based on GeP, and two groups of control experiments (e$\rightarrow$f and g$\rightarrow$h) are conducted. Comparing with c$\rightarrow$e, c$\rightarrow$f proves that our method can solve the difficulty of confidence learning in the projection-based model. 
The experimental results clearly demonstrate the effectiveness of our geometry modeling method for all metrics.

  \begin{table}[!t]
    \centering
    \fontsize{7}{10}\selectfont
    \caption{Comparison with combinations of our GUP Net with some other widely used loss weights controllers on the KITTI \emph{validation} set for the car category.}
    \label{tab:weight_comp}
    \begin{tabular}{M{3cm}| M{0.3cm} M{0.3cm} M{0.3cm} M{0.3cm} M{0.3cm} M{0.3cm}}
        \toprule
        \multirow{2}{*}{Loss weight controller}&
        \multicolumn{3}{c}{3D@IoU=0.7}&\multicolumn{3}{c}{BEV@ IoU=0.7}\cr\cline{2-7}
        &Easy&Mod.&Hard&Easy&Mod.&Hard\cr\hline
        GradNorm~\cite{chen2018gradnorm}&16.19&10.49&9.04&21.80&14.74&13.02\cr
        Task Uncertainty~\cite{kendall2018multi}&18.95&13.94&12.18&25.07&19.45&16.74\cr
        HTL (Ours)&\textbf{22.76}&\textbf{16.46}&\textbf{13.72}&\textbf{31.07}&\textbf{22.94}&\textbf{19.75}\cr
        \bottomrule        
    \end{tabular}
\end{table}
\noindent
{\bf Influence of the Hierarchical Task Learning}.
We also quantify the contribution of the proposed Hierarchical Task Learning (HTL) strategy by two groups of control experiments (e$\rightarrow$g and f$\rightarrow$h), and both of them confirm the efficacy of the proposed HTL (improving performances for all metrics, and about 2\% improvements for easy level).
Also, we investigate the relationships between the loss terms and visualize the changing trend of loss weights in the training phase in Figure~\ref{fig:loss_weight} to indicate the design effectiveness of our HTL scheme. It shows that the 2nd stage loss weights start increasing after all its tasks (\{heatmap, 2D offset and 2D size\}) close to convergence. And for the 3rd depth inference stage, it has a similar trend. Its loss weight starts increasing at about 11th epochs. At that time, all its pre-tasks \{heatmap, 2D offset and 2D size, 3D size\} have achieved certain progress. 

To further prove that this strategy fits our method, we also compare our HTL with some widely used loss weight controllers~\cite{chen2018gradnorm,kendall2018multi} in Table~\ref{tab:weight_comp}. We can see that our methods achieve the best performance. The main reason for the poor performance of the comparison methods is that our model is a hierarchical task structure. The task-independent assumption they request does not hold in our model. And for the GardNorm, its low performance is also caused by the error amplification effect. This effect makes the magnitude of the loss function significantly change in the total training phase so it is hard for the 
GardNorm to balance them. 

\subsection{Qualitative Results}
For further investigating the effectiveness of our GUP Net. 
We show some bad cases and corresponding uncertainties from our model and the baseline projection method (the same setting in the 4th line in Table~\ref{tab:ablation}). The results are shown in Figure~\ref{fig:uncertain}. We can see that our GUP Net can predict with high uncertainties for different bad cases including occlusion and far distance. And with the improvement of the prediction results, the uncertainty prediction of our method basically decreases. And the baseline projection model gives similar low uncertainty values for that bad case, which demonstrates the efficiency of our GUP Net.

\section{Conclusion}
In this paper, we proposed GUP Net model for the monocular 3D object detection to tackle the error amplification ignored by conventionally geometry projection models. It combines mathematical projection priors and the deep regression power together to compute more reliable uncertainty for each object, which not only be helpful for the uncertainty-based learning but also can be used to compute the accurate confidence in the testing stage. We also proposed a Hierarchical Task Learning strategy to learn the overall model better and reduce the instability caused by the error amplification. Extensive experiments validate the superior performance of the proposed algorithm, as well as the effectiveness of each component of the model.

\section{Acknowledgement}
This work was supported by the Australian Research Council Grant DP200103223, FT210100228, and Australian Medical Research Future Fund MRFAI000085.

\end{document}